\documentclass[11pt]{article}
\usepackage[preprint]{acl}
\usepackage{times,latexsym}
\usepackage[T1]{fontenc}
\usepackage[utf8]{inputenc}
\usepackage{microtype,courier}
\usepackage{graphicx,booktabs,amsmath,amssymb,tabularx}
\usepackage{xurl}

\newcommand{\C}{\mathrm{C}}
\newcommand{\A}{\mathrm{A}}

\title{The Concrete--Arbitrary Gap: Kinship Reasoning in LLMs Is Not Indifferent to Presentation}
\author{Thomas Pashby\\
Middlebury College\\
\texttt{tpashby@middlebury.edu}}
\hypersetup{
  pdftitle={The Concrete-Arbitrary Gap: Kinship Reasoning in LLMs Is Not Indifferent to Presentation},
  pdfsubject={Presentation dependence in relational reasoning by large language models},
  pdfkeywords={large language models, relational reasoning, kinship inference, semantic content, reasoning budgets, representation sensitivity}}

\begin{document}
\maketitle
\begin{abstract}
We test whether large language models solve formally matched kinship problems equally well when relations are expressed in familiar vocabulary or by explicitly defined nonce predicates. Across 500 paired graphs, concrete accuracy exceeds arbitrary accuracy by 35.6 percentage points in local Qwen3.8-27B, 26.6 in Gemma 4 26B-A4B, 12.0 in Gemma 4 31B, and 5.4 in Qwen3.8-Max. All four paired gaps are statistically resolved. Reasoning budgets and prompt-language interventions can substantially reduce the difference, showing that it is modifiable rather than a fixed incapacity. The minimal conclusion is behavioral: on these tasks, the models’ manifested relational competence is not indifferent to presentation. Explicit definitions provide the formal relations but do not make nonce predicates as usable as familiar vocabulary embedded in learned linguistic associations.
\end{abstract}

\section{Introduction}
Relational reasoning is often tested by asking a model to follow a chain of stated facts. Consider a reasoner whose competence is indifferent to the representation of those facts. Once told that \texttt{wug(x,y)} means that $x$ and $y$ are siblings and \texttt{nuk(x,y)} means that $x$ is a parent of $y$, it should be able to compose \texttt{wug(Julia,Ben)} and \texttt{nuk(Ben,Eric)} just as it composes ``Julia and Ben are siblings; Ben is Eric's parent.'' The symbols differ, but the stipulated relations and required composition do not.

This yields a direct behavioral prediction. Holding the graph, query, and answer fixed, a presentation-indifferent competence should not systematically succeed more often in one vocabulary than the other. Equality would not by itself prove an invariant internal algorithm, but a reliable paired difference is enough to reject presentation-indifference for the competence manifested in the task. The relevant contrast is not between meaningful and undefined symbols. The nonce predicates are explicitly defined. It is between familiar expressions that participate in a rich network of learned linguistic and inferential associations and predicates whose task-relevant content is supplied by a local stipulation.

We examine this difference on 500 paired synthetic kinship graphs. Each graph has a concrete rendering with ordinary relation terms and an arbitrary rendering with nonce predicates whose meanings are explicitly defined. The queried people and correct English answer are the same. The task is not memorization of a particular graph, nor does it ask the model to infer the definitions of unknown signs. It asks whether a known relational composition transfers across two ways of stating the same facts.

The central result is a \emph{concrete--arbitrary gap}: all four tested models answer more of the concrete prompts correctly when giving an answer without a visible intermediate analysis. The gap is large in the two locally hosted models and remains detectable in a hosted model that is near ceiling. Thus, for these models and problems, relational competence is not indifferent to presentation. This conclusion concerns the \emph{whole} process of reading, composing, and answering. It does not entail that every latent operation is content-dependent: an invariant composition procedure downstream of a presentation-sensitive interpretation stage remains possible.

We also ask how stable the gap is. Providing a dedicated reasoning budget often reduces it; displaying the set of English relation terms or requesting a coded answer can alter it even without a thinking channel. Those manipulations are valuable precisely because they show that the gap is not an immutable incapacity. They do not make its initial occurrence less real. The organizing question of this paper is therefore \emph{where and under what presentation the gap appears}, with its repair treated as evidence about the conditions of access to relational competence.

\section{Related Work}
Large language models show analogical performance on novel tasks \citep{webb2023}, complicating a simple recollection account. Yet linguistic and semantic content affects reasoning \citep{lampinen2024}; counterfactual task variants challenge unrestricted transfer \citep{wu2024,lewis2024}; and the encoding of formally similar graph information matters \citep{fatemi2024}. A paired presentation test isolates a particularly clear case: the relations are defined, the graph is held fixed, and the output question is shared.

Additional inference-time computation can improve model answers \citep{wei2022}, but a generated reasoning trace is not a transparent record of the hidden computation \citep{turpin2023,jacovi2020}. We consequently treat reasoning budget as an operational intervention and answer-channel elaboration as a protocol feature. Neither a native ``thinking off'' switch nor an empty separate reasoning field alone establishes that a response was answer-only.

\section{Task and Evaluation}
\subsection{Paired kinship graphs}
The redesigned-v3 dataset contains 500 generated graphs, each with concrete and arbitrary prompts. Fifteen target relations span grandchild, avuncular, niece/nephew, cousin, great-avuncular, and marriage-linked families. Primitive facts concern directed parenthood, symmetric siblinghood and marriage, and gender; the benchmark adopts deliberately simplified conventions. Target classes are nearly balanced. The preidentified 434-graph subset excluding great-aunt/uncle is a sensitivity analysis, not a replacement for the full set.

An actual matched pair (graph \texttt{kn-0062}) makes the manipulation plain. The concrete prompt states: ``Julia is female. Ben is male. Eric is male. Ben is a parent of Eric. Julia and Ben are siblings. What is Julia's relationship to Eric?'' The arbitrary prompt states: ``\texttt{nuk(x,y)} means x is a parent of y; \texttt{wug(x,y)} means x and y are siblings; \texttt{tor(x,y)} means x and y are married. \texttt{nuk(Ben,Eric)}, \texttt{wug(Julia,Ben)},'' together with the same gender facts and question. Both require \emph{aunt}. The third, unused predicate definition is included by the task renderer, not introduced only in the arbitrary answer.

The pairing fixes the underlying graph, named query, primitive denotations, and gold relation. It does not hold surface length, syntax, or every fact position fixed: definitions lengthen arbitrary prompts, and the two renderings can shuffle facts independently. The reported effect is therefore the effect of this \emph{presentation package}, not a pure substitution of one token string for another. Crucially, the arbitrary condition supplies ordinary-English definitions. The model need not infer the predicates from examples; it must use their stipulated meanings in the required composition.

We operationalize presentation-indifference by the minimal null hypothesis
\begin{equation}
H_{\mathrm{PI}}:\quad \mathbb{E}_{g}[Y_g^\C-Y_g^\A]=0,
\end{equation}
where the expectation is over matched graphs. This is a claim about successful performance by the complete system under the two explicit presentations. It is weaker than a claim that all processing stages must be identical and stronger than the claim that some downstream operation could, in principle, be invariant after successful interpretation.

\subsection{Models and answer-only protocols}
We test locally hosted Qwen3.8-27B-AWQ-INT4 through vLLM 0.29.0 \citep{qwen3827b2026,qwenlocal2026,vllm}, Gemma 4 26B-A4B and dense Gemma 4 31B through llama.cpp b10917/Vulkan \citep{gemma426b2026,gemma431b2026,llamacpp}, and hosted Qwen3.8-Max \citep{qwenmax2026}. The local device is an RTX 3090. Temperature is 1.0, top-$p$ 0.95, top-$k$ 20, and the numerical seed label is 20260914 in the main cells. Shared seed labels do not imply shared random draws across backends.

The lead result uses \emph{observed answer-only} outputs. For local Qwen and Gemma 26B, we ran new strict whole-response cells: the dedicated reasoning channel was disabled, the instruction requested only one canonical relation, and the completion allowance was 16 tokens. The entire response---not just its final line---was scored. The 1,000 outputs in each cell had no multiline answer, separate reasoning trace, or truncation. Gemma 31B's repaired off cell and Qwen-Max's off cell retained their original, more generous envelopes and final-line scoring, but all recorded responses were single-line and had no separate reasoning trace. These are two \emph{different} protocol types, displayed separately in the lead figure. A single-word answer does not show an absence of internal computation; it only rules out visible answer-channel deliberation.

For each graph $g$ let $Y_g^\C$ and $Y_g^\A$ indicate conservative correctness. The paired gap is
\begin{equation}
G=\frac{1}{500}\sum_{g=1}^{500}(Y_g^\C-Y_g^\A).
\end{equation}
Invalid, noncanonical, truncated, and request-error answers count as failures. We give two-sided exact McNemar tests and 10,000-resample graph-bootstrap 95\% intervals. Graphs, not 1,000 separate responses, are the sampling units. These intervals describe variation over generated graphs and not over seeds, model checkpoints, or independent datasets.

\section{The Concrete--Arbitrary Gap}
\begin{figure*}[t]
\centering\includegraphics[width=.89\textwidth]{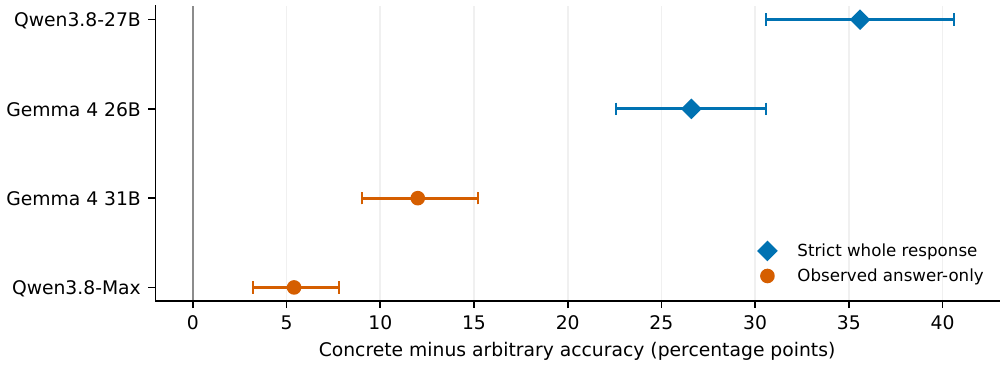}
\caption{Paired concrete-minus-arbitrary accuracy on the same 500 graphs per model. Diamonds denote strict whole-response reruns; circles denote original off cells whose saved outputs were observationally answer-only. Intervals resample graphs. Marker types flag distinct protocols, not model architectures.}
\label{fig:headline}
\end{figure*}

Figure~\ref{fig:headline} shows a positive gap in every model. Table~\ref{tab:headline} gives the component accuracies, uncertainty, and paired discordances. Local Qwen has 203 graphs solved only in concrete form against 25 solved only in arbitrary form; Gemma 26B has 135 against 2. Even Qwen-Max, at 93.0\% arbitrary accuracy, has 32 concrete-only against 5 arbitrary-only successes. Exact McNemar $p$-values are respectively $7.71\times10^{-36}$, $1.09\times10^{-37}$, $5.87\times10^{-15}$, and $7.43\times10^{-6}$. A four-test Holm correction does not change the inference that each paired gap differs from zero.

\begin{table*}[t]
\centering\small\setlength{\tabcolsep}{5pt}
\begin{tabular}{lrrrrrr}\toprule
Model & Concrete & Arbitrary & Gap & Gap 95\% CI & C-only/A-only & Valid C/A\\\midrule
Qwen3.8-27B & 64.6 & 29.0 & +35.6 & [30.6,40.6] & 203/25 & 394/253 \\
Gemma 4 26B & 96.8 & 70.2 & +26.6 & [22.6,30.6] & 135/2 & 499/496 \\
Gemma 4 31B & 98.2 & 86.2 & +12.0 & [9.0,15.2] & 64/4 & 496/491 \\
Qwen3.8-Max & 98.4 & 93.0 & +5.4 & [3.2,7.8] & 32/5 & 496/495 \\
\bottomrule\end{tabular}
\caption{Answer-only baseline percentages (gap in points). ``Valid'' gives canonical responses out of 500 in each presentation. Invalid outputs remain incorrect. Local Qwen and Gemma 26B use strict whole-response scoring; Gemma 31B and Qwen-Max use original final-line scoring, but every saved output is single-line.}
\label{tab:headline}
\end{table*}

Output validity is especially important for local Qwen: only 394/500 concrete and 253/500 arbitrary answers are canonical. Many invalids are short kinship words outside this task's accepted ontology, such as \emph{sister}, \emph{father}, or the gender-neutral \emph{grandchild}; six concrete outputs are empty. They are not all failures of relational composition. Conditional accuracy among valid outputs is 82.0\% concrete and 57.3\% arbitrary, but conditioning selects different subsets and is not a replacement endpoint. The presentation effect is therefore a measured difference in successful \emph{canonical task performance}; its allocation among interpretation, composition, and label selection remains open. The near-universal validity for the other three models prevents this Qwen-specific qualification from explaining away the four-model result.

\begin{table}[t]
\centering\small\setlength{\tabcolsep}{3pt}
\begin{tabular}{lrrrr}\toprule
Relation family ($n$) & Q27 & G26 & G31 & Q-Max\\\midrule
Grandchild (68) & +16.2 & +2.9 & +2.9 & +1.5 \\
Lateral (168) & +41.7 & +19.6 & +17.3 & +0.0 \\
Great-avuncular (66) & -4.5 & +51.5 & +31.8 & +31.8 \\
Marriage-linked (198) & +50.5 & +32.3 & +4.0 & +2.5 \\
\bottomrule\end{tabular}
\caption{Answer-only C$-$A gap in points by gold relation family. Family counts sum to 500. These descriptive partitions show heterogeneity and are not four independent confirmatory tests per model.}
\label{tab:families}
\end{table}

Table~\ref{tab:families} shows that the gap is not distributed uniformly. It is large for local Qwen on lateral and marriage-linked relations, while its great-avuncular concrete accuracy is at floor. Gemma 26B has sizeable gaps in lateral, great-avuncular, and marriage-linked cases. Dense Gemma retains a positive gap in those families. Qwen-Max's small overall gap is concentrated in great-aunt/uncle: excluding those 66 graphs leaves 428/434 concrete and 422/434 arbitrary successes, a 1.4-point difference (10 concrete-only versus 4 arbitrary-only, exact $p=0.180$). The full-set result is valid, but it would mislead to describe the hosted model as broadly presentation-sensitive across every relation family. The other three models retain substantial gaps on this predefined 434-graph subset: 41.7, 22.8, and 9.0 points respectively.

The original local Qwen and Gemma 26B ``thinking off'' cells yielded gaps of 32.4 and 15.4 points. About 40\% of those cells' answers contained multiple lines, despite lacking a separate reasoning field. The new answer-only gaps are 35.6 and 26.6. This corroborates the direction of presentation dependence under a protocol that blocks visible answer-channel scratchpads. It is \emph{not} a clean estimate of suppressing deliberation: the reruns changed the instruction and output limit, and the strict-minus-original gap change for local Qwen is only 3.2 points with a graph-bootstrap interval from $-3.2$ to $9.6$. The answer-only cells do not support an unqualified cross-model ranking because they use two protocol types.

\subsection{What a paired failure looks like}
Graph \texttt{kn-0072} is an illustrative case, not a selected statistical endpoint. In concrete form it says that Adam and Kevin are married, Iris is Kevin's parent, and Adam is male; it asks for Adam's relation to Iris. All four answer-only cells give the canonical \emph{son-in-law}. In arbitrary form, the same facts are \texttt{fep(Adam,Kevin)} and \texttt{wug(Iris,Kevin)}, where \texttt{fep} is explicitly defined as marriage and \texttt{wug} as directed parenthood. The query and gold relation are unchanged. Local Qwen answers \emph{husband}, Gemma 26B \emph{brother-in-law}, Gemma 31B \emph{husband}, and Qwen-Max \emph{husband}. These short outputs illustrate distinct errors that the aggregate correctness measure combines. \emph{Husband} follows one primitive fact but answers a relation to Kevin rather than the stated relation to Iris; \emph{brother-in-law} selects the wrong composition. No inference about typical error type follows from one selected graph, but the case shows that the arbitrary prompt's explicit definitions do not force correct use of the query perspective.

\subsection{What the paired design does and does not control}
Within a graph, the gold answer, entity names, primitive relation meanings, and queried direction are fixed. This removes several mundane explanations for a difference in accuracy: a different target class, a different graph, or a different answer vocabulary cannot explain the paired contrast. The exact McNemar test uses only discordant pairs, rather than treating the two presentations as unrelated sets of 500 observations. Conversely, pairing does not remove the processing costs of definition lookup, symbolic argument order, prompt length, or reordered facts. These are plausible pathways by which presentation acts. The result is intentionally broad: a model's ability to solve the \emph{posed} relational problem is not invariant under this explicit recoding.

One might object that a sufficiently capable system could first translate arbitrary predicates into familiar terms and then use the same composition procedure. We agree that this architecture is compatible with the result. But it does not rescue presentation-indifference of the manifested competence: the translation is part of solving the arbitrary problem, and the measured gaps show that the complete system does not perform it reliably enough to preserve success across vocabularies. A narrower hypothesis---that composition is invariant \emph{conditional on successful interpretation or translation}---requires direct intermediate-state or controlled-intervention evidence and is not rejected here.

\section{How the Gap Changes}
\subsection{Reasoning budgets}
\begin{figure*}[t]
\centering\includegraphics[width=.96\textwidth]{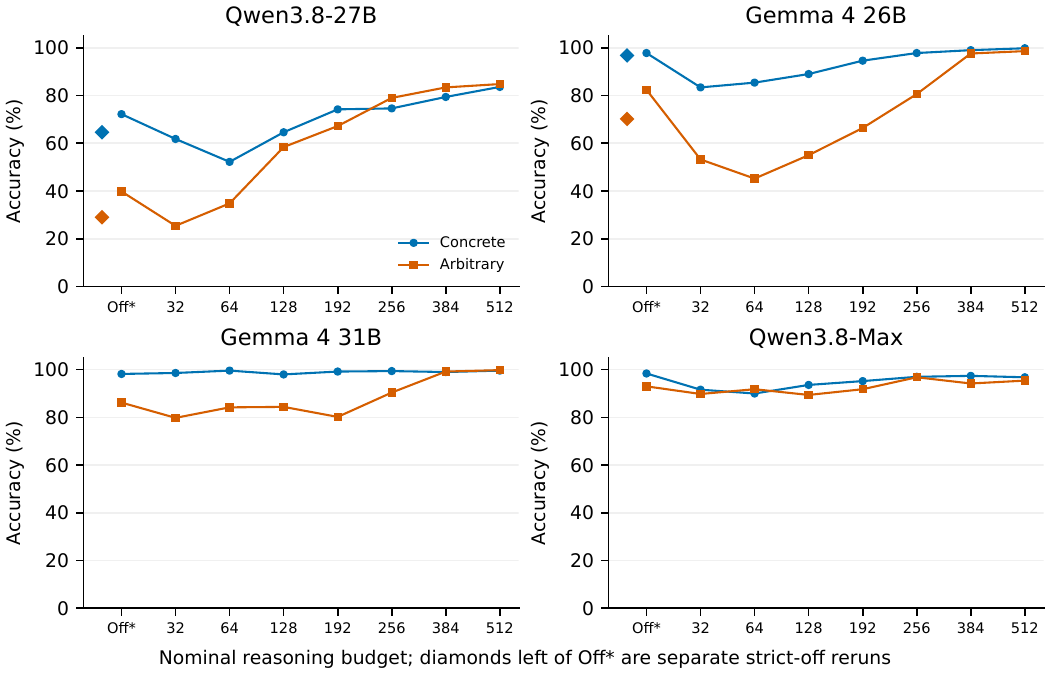}
\caption{Original common-protocol accuracy curves (500 graph pairs per point). Off* means the dedicated reasoning channel was disabled, but local Qwen and Gemma 26B sometimes elaborated in the answer channel. Unconnected diamonds to its left show their separate strict answer-only reruns; no line joins them to the budget curve. Horizontal positions are categorical.}
\label{fig:budget}
\end{figure*}

The original common-protocol grid varies nominal dedicated-channel budgets at off, 32, 64, 128, 192, 256, 384, and 512 tokens. Figure~\ref{fig:budget} plots accuracy within each model. The connected off points mean \emph{reasoning channel disabled}; the strict local reruns appear as unconnected diamonds. The serving systems implement budget transitions differently, and actual reasoning-token counts are not a common compute unit.

Sufficient budget substantially reduces the original off gap in the three local models. At 512, local Qwen scores 83.6\% concrete and 84.8\% arbitrary; Gemma 26B scores 99.8\% and 98.6\%; Gemma 31B scores 99.6\% and 99.8\%. These small differences are not equivalence results. More importantly, closure is not merely a loss of concrete accuracy: arbitrary accuracy rises by 45.0, 16.2, and 13.6 points relative to each model's \emph{original common-protocol off} cell. Local Qwen's concrete accuracy rises by 11.4 points, Gemma 26B's by 2.0, and Gemma 31B's by 1.4.

Absolute performance comparisons are also informative when kept benchmark-specific. At the nominal 512-token setting, Gemma 31B exceeds Qwen-Max by 2.8 points on concrete prompts (99.6\% versus 96.8\%; 16 Gemma-only successes versus 2 Qwen-only; exact McNemar $p=.00131$) and by 4.4 points on arbitrary prompts (99.8\% versus 95.4\%; 23 versus 1; $p=2.98\times10^{-6}$). Graph-bootstrap intervals are 1.2--4.6 and 2.6--6.4 points, respectively; averaging the two presentation outcomes within graph gives a 3.6-point advantage (interval 2.3--5.1). Thus Gemma 31B performs better on this task under these recorded settings. The comparison does not establish generally superior reasoning: the models differ in training, architecture, serving backend, and implementation of a nominal reasoning budget, and this comparison uses one decoding seed.

Short budgets need not help. At 32, arbitrary accuracy is lower than original off in all four models, and Gemma 26B falls further at 64. The pattern is not universal in its shape: Qwen-Max begins close to ceiling, varies within a narrower range, and its 512 result (96.8\% concrete, 95.4\% arbitrary) combines a 2.4-point arbitrary gain with a 1.6-point concrete decline relative to off. We therefore present the budget curves as a study of the gap's \emph{contingency}, not as a single monotonic law or a cross-model measure of equivalent reasoning effort.

The curve's first step deserves care. ``Off'' disables a channel; ``32'' imposes a short separate trace and a subsequent transition to the answer. For local Qwen and Gemma 26B, the off answer channel was often long enough to contain visible working; at 32, the model may be forced out of its dedicated trace before that working is complete. A lower 32-token score need not mean that a small amount of genuine inference is harmful. It could reflect the interaction of a short budget, forced termination, and final-answer compliance. The stronger within-model finding is that presentation-specific success changes substantially across operational settings. The present records do not identify one universal computational threshold.

\subsection{Prompt language and output vocabulary}
A separate, frozen follow-up adds isomorphic distractor components to the graphs and varies output instructions for Gemma 31B and local Qwen at thinking off. The \emph{free} condition requests an English relation without a displayed answer inventory. The \emph{inventory} condition prints the 15 permissible English labels. The \emph{coded} condition maps those labels to graph-specific letters and requests the corresponding letter. These are paired \emph{within this augmented dataset}; their accuracies should not be numerically compared with the original budget grid.

\begin{table}[t]
\centering\small\setlength{\tabcolsep}{4pt}
\begin{tabular}{llrrr}\toprule
Model & Output & C & A & Gap\\\midrule
Qwen3.8-27B & Free & 61.2 & 49.6 & +11.6 \\
Qwen3.8-27B & Inventory & 78.8 & 88.6 & -9.8 \\
Qwen3.8-27B & Coded & 87.2 & 89.0 & -1.8 \\
Gemma 4 31B & Free & 92.0 & 56.4 & +35.6 \\
Gemma 4 31B & Inventory & 95.6 & 91.6 & +4.0 \\
Gemma 4 31B & Coded & 99.4 & 99.0 & +0.4 \\
\bottomrule\end{tabular}
\caption{Augmented 500-graph follow-up, percentages and C$-$A points. ``Inventory'' displays English relation labels; ``coded'' displays a randomized label-to-letter mapping. This is a different task set and protocol from Figure~\ref{fig:budget}.}
\label{tab:prompt}
\end{table}

Gemma 31B's free condition has a 35.6-point gap, reduced to 4.0 with an inventory and 0.4 with coded output. Its free-to-inventory gap change is 31.6 points (paired graph-bootstrap interval 27.0--36.4). Local Qwen's 11.6-point free gap changes to $-9.8$ with an inventory and $-1.8$ with coded output; its free-to-inventory change is 21.4 points (interval 14.2--28.6). Thus the prompt can reduce or reverse the gap; the models do not respond to the manipulation identically. The inventory supplies familiar answer vocabulary, while coding also changes final-answer production and compliance. Neither manipulation isolates translation as a necessary inner step. Their importance here is that a robust presentation effect is also \emph{instruction-sensitive}.

The prompt follow-up guards against treating the original gap as a permanent ranking of concrete over arbitrary expressions. It also blocks an overly easy conclusion that a displayed inventory alone makes a model operate over nonce predicates exactly as it does over kinship terms: the full prompt changes, the output candidate set becomes explicit, and the coded condition adds a lookup operation. Local Qwen even reverses the sign of the gap with the inventory, whereas dense Gemma retains a smaller concrete advantage. The two models' different responses are evidence against a one-size-fits-all account of ``semantic scaffolding.'' They support the narrower point that answer vocabulary and task instructions are experimentally relevant to the availability of relational competence.

\section{Discussion and Conclusion}
Across four models, the same relational graph is more likely to be answered correctly when its primitives are rendered in ordinary kinship language. Because the arbitrary predicates are explicitly defined, this is not a comparison between known relations and uninterpreted symbols. The formal information required for the inference is supplied in both conditions. What differs is whether that information arrives in vocabulary already situated within the model's learned linguistic network or through local definitions that must be put to use. Explicit definition is therefore not sufficient for equal performance.

The rejected hypothesis is $H_{\mathrm{PI}}$: equal concrete and arbitrary success by the full presented-task pipeline on this graph distribution. The conclusion is correspondingly direct: the relational competence manifested by these models on kinship problems is not indifferent to presentation. A stronger architectural thesis---that the models contain no invariant relational operation anywhere---is not tested by the accuracy gap. Such an operation might occur after fragile parsing or translation, or alongside learned semantic routines. That possibility relocates rather than removes the behavioral dependence: a reasoner that fails to make formally supplied relations available to its putative solver does not display presentation-indifferent competence on the posed task.

This gives a limited connection to material inference \citep{brandom1994}. The transition from ``female sibling of a parent'' to \emph{aunt} is licensed by the content of the terms, not by any formal rule. Familiar kinship expressions have what we may call greater \emph{semantic valence}: they stand in many learned linguistic and inferential relations that a one-sentence nonce definition does not recreate. The observed advantage for familiar vocabulary is consistent with models exploiting those semantic relations. Budget and inventory effects make familiar vocabulary a plausible scaffold but the data do not establish its causal necessity. The obvious intervention of instructing a model not to translate into familiar language was attempted but unsuccessful since the models ignored this instruction.

The results also make a positive methodological case for studying accessible local models. Gemma 4 31B, running locally on a single RTX 3090, outperforms the hosted flagship model Qwen3.8-Max \citep{qwenmax2026} at the nominal 512-token setting on both presentations of this benchmark. This is not a general model ranking, but it shows that local models need not be scientifically interesting only as cheap approximations to stronger services. For a targeted linguistic phenomenon, they can be the better-performing experimental system.

Local control also changes what can be measured. We could inspect answer-channel behavior, diagnose a flawed ``off'' baseline, and rerun the identical graphs under a strict protocol without replacing earlier artifacts. Open, locally hosted models additionally permit interventions unavailable through an API: recording layer tensors, examining routing and activation trajectories, replaying exact token sequences, and modifying inference code. Such access makes mechanistic data available without renting or operating a server rack for every experiment. A single consumer GPU still consumes energy, and the upstream cost of training the checkpoint is not removed; the narrower point is that inference, replication, and tensor-level observation can be conducted with substantially less financial and environmental overhead than large multi-GPU deployments. This combination of competitive task performance, experimental control, and modest infrastructure makes small and locally deployable models valuable objects of linguistic study in their own right.

The interpretation should nevertheless respect model scale and ceiling. The hosted Qwen-Max gap is small in absolute terms, and the family analysis places much of it in the great-avuncular category. It provides a clear paired difference on the full frozen set, while the larger local gaps make the phenomenon easier to observe and audit. These results are not estimates of a universal law of scale. They establish that parameter scale and deployment status do not determine performance on this task, and that the presentation effect is not confined to a single model or to invalid formatting in one local checkpoint. The next test should separate comprehension of primitive definitions, relational composition, and canonical-label production under independently controlled output protocols. The present contribution is the reproducible gap that such an account must explain.

\label{mainend}
\clearpage
\section*{Limitations}
The benchmark is synthetic and its 500 graphs instantiate a small family of short chains. Bootstrap intervals describe variation within that family, not general relational reasoning. The arbitrary rendering adds definitions, changes length and syntax, and sometimes changes fact order. The study does not locate which of these properties creates the difference. The answer is always an English kinship term in the original task; even a correct nonce-language solution must select a familiar label. The follow-up with coded outputs reduces but does not eliminate reliance on an English codebook.

The four models have different architectures, checkpoints, quantization, training histories, and serving systems. Their answer-only baselines are observationally comparable as short responses, not identical interventions. The strict local reruns changed both instruction and completion cap. The original budget grids are internally paired but their off cells do not always prevent visible answer-channel elaboration. Nominal budget values are not common amounts of computation, and most conditions have only one decoding seed. The Gemma 31B--Qwen-Max comparison at 512 is therefore a benchmark-and-protocol result, not a controlled estimate of architecture, parameter count, or general capability. We neither claim equivalence from a small or nonsignificant gap nor infer a pure effect of hidden reasoning from visible traces.

Noncanonical outputs are counted as failures and occur especially often for local Qwen; valid-only sensitivity conditions on a post-treatment variable. High Qwen-Max accuracy leaves less headroom for a large absolute gap. The semantic-inventory follow-up uses a harder augmented graph set and different instructions, so its absolute scores cannot calibrate the original budget curve. Earlier exploratory development informed task design; the entire research program was not prospectively fixed. No internal-mechanism claim is made from routing, activation probes, or trace language in this paper.

\section*{Ethical Considerations}
All names and family graphs are synthetic. The task uses simplified binary gender labels and idealized marriage and kinship relations; these are benchmark conventions, not a description of the diversity of real families. Local GPU runs and hosted requests use computational resources, but we have not measured their complete energy cost. Model use remains subject to checkpoint licenses and API terms.

\section*{Data and Code Availability}

The frozen prompts, model responses, machine-readable results, evidence ledger, and analysis code are available at \url{https://github.com/tompashby/concrete-arbitrary-gap}.

\bibliography{reasoning_portability_acl}

\end{document}